\documentclass[10pt,twocolumn,letterpaper]{article}

\usepackage[pagenumbers]{cvpr} 
\usepackage{makecell}
\usepackage{multirow}
\usepackage{arydshln}
\usepackage{booktabs}
\usepackage[listings,theorems]{tcolorbox}

\definecolor{cvprblue}{rgb}{0.21,0.49,0.74}
\usepackage[pagebackref,breaklinks,colorlinks,allcolors=cvprblue]{hyperref}

\def\paperID{1699} 
\def\confName{CVPR}
\def\confYear{2025}

\title{PEACE: Empowering Geologic Map Holistic Understanding with MLLMs}

\author{\textbf{Yangyu Huang}$^{1}$\thanks{Equal contribution}\;\;\;\textbf{Tianyi Gao}$^{1*}$\;\;\;\textbf{Haoran Xu}$^{1}$\;\;\;\textbf{Qihao Zhao}$^{1}$\;\;\textbf{Yang Song}$^{2}$\;\;\;\\\textbf{Zhipeng Gui}$^{3}$\;\;\;\textbf{Tengchao Lv}$^{1}$\;\;\;\textbf{Hao Chen}$^{1}$\;\;\;\textbf{Lei Cui}$^{1}$\;\;\;\textbf{Scarlett Li}$^{1}$\thanks{Corresponding author}\;\;\;\textbf{Furu Wei}$^{1}$\\
{$^{1}$Microsoft Research} \;\;\;\;
{$^{2}$Chinese Academy of Geological Sciences} \;\;\;\;
{$^{3}$Wuhan University}\\
{\tt\small \{yanghuan,v-tianyigao,v-haoraxu,v-qihaozhao,tengchaolv,chhao,lecu,scarli,fuwei\}@microsoft.com}\\
{\tt\small song.yang@cags.ac.cn\;\;\;\;zhipeng.gui@whu.edu.cn}
}

\begin{document}
\maketitle
\begin{abstract}

Geologic map, as a fundamental diagram in geology science, provides critical insights into the structure and composition of Earth's subsurface and surface.
These maps are indispensable in various fields, including disaster detection, resource exploration, and civil engineering.
Despite their significance, current Multimodal Large Language Models (MLLMs) often fall short in geologic map understanding.
This gap is primarily due to the challenging nature of cartographic generalization, which involves handling high-resolution map, managing multiple associated components, and requiring domain-specific knowledge.
To quantify this gap, we construct \textbf{GeoMap-Bench}, the first-ever benchmark for evaluating MLLMs in geologic map understanding, which assesses the full-scale abilities in extracting, referring, grounding, reasoning, and analyzing.
To bridge this gap, we introduce \textbf{GeoMap-Agent}, the inaugural agent designed for geologic map understanding,
which features three modules: Hierarchical Information Extraction (HIE), Domain Knowledge Injection (DKI), and Prompt-enhanced Question Answering (PEQA).
Inspired by the interdisciplinary collaboration among human scientists, an AI expert group acts as consultants, utilizing a diverse tool pool to comprehensively analyze questions.
Through comprehensive experiments, GeoMap-Agent achieves an overall score of 0.811 on GeoMap-Bench, significantly outperforming 0.369 of GPT-4o.
Our work, em\textbf{P}owering g\textbf{E}ologic m\textbf{A}p holisti\textbf{C} und\textbf{E}rstanding (\textbf{PEACE}) with MLLMs, paves the way for advanced AI applications in geology, enhancing the efficiency and accuracy of geological investigations.
\end{abstract}
\vspace{-10px}
    
\section{Introduction}
\label{sec:intro}


\begin{figure}
  \centering
  \includegraphics[width=3.3in]{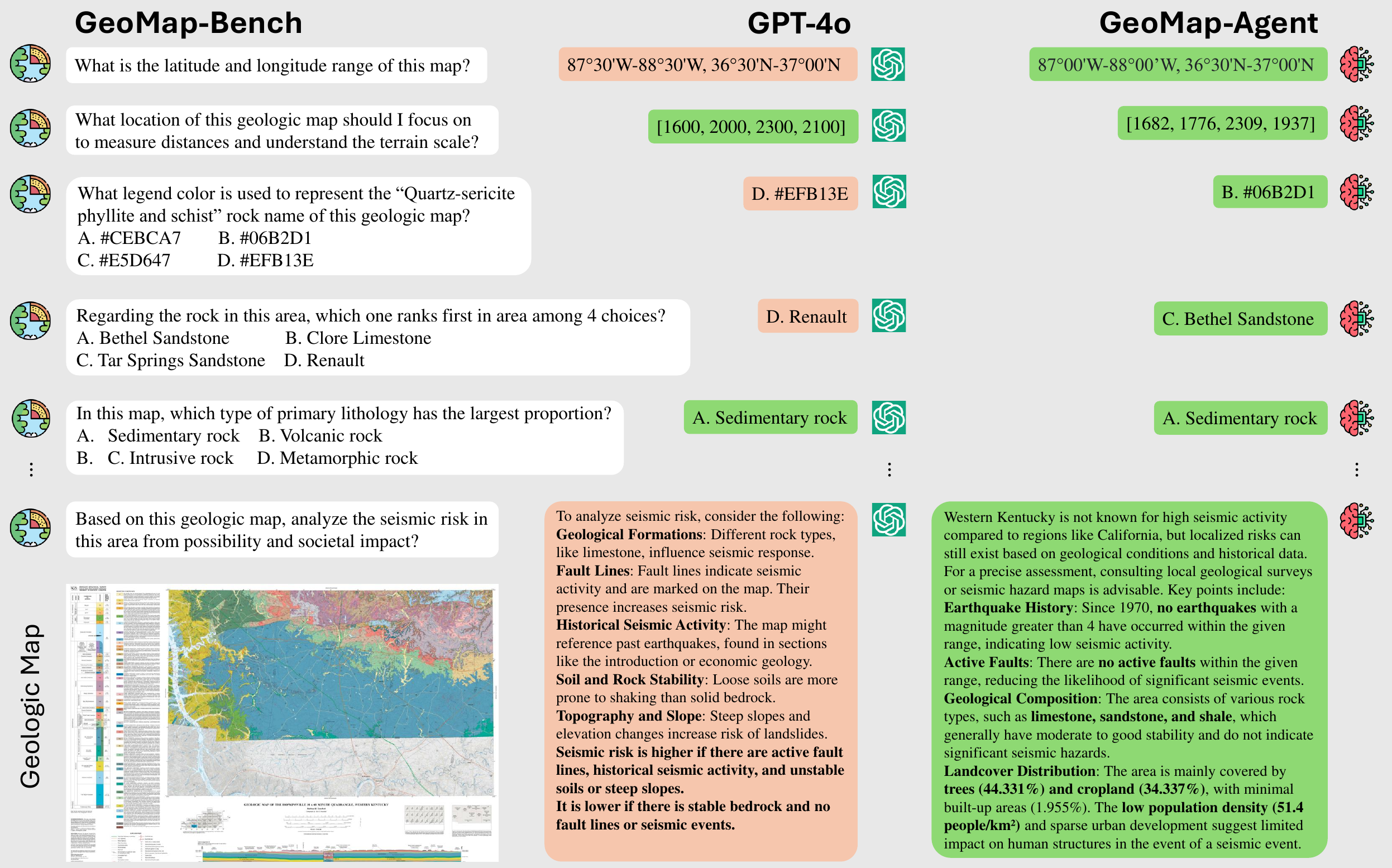}
  \hfill
\vspace{-10px}
  \caption{The sampled questions of GeoMap-Bench and the corresponding answers by GeoMap-Agent and GPT-4o respectively.
  Green chatboxes indicate correct answers, while red ones denote incorrect or poor.
  GeoMap-Agent provides accurate answers for basic questions and comprehensive responses for advanced ones.
  GPT-4o, although able to correctly answer some basic questions, struggles with advanced ones.
  For instance, the response to the last question is too vague to provide concrete information.
  }
  \label{fig:benchmark_construction}
\vspace{-10px}
\end{figure}


Geology science plays a pivotal role in comprehensively understanding the Earth, offering insights into the processes that have shaped our planet over 4.5 billion years.
The Earth's history is divided into several eons, eras, periods, and epochs, each characterized by significant geological and biological events.
Geologic map is crucial in deciphering this history, as it documents the distribution and relationships of rock units and geological features.


As for application, geologic map plays a key part in the following scenarios:
(1) \textbf{Disaster Detection}: assessing risks of natural hazards like landslides, earthquakes, and groundwater contamination.
(2) \textbf{Resource Exploration}: identifying potential locations for natural resources such as mining minerals, oil, and gas.
(3) \textbf{Civil Engineering}: planning infrastructure projects by understanding the subsurface conditions.
For instance, seismic activity is closely linked to active fault systems in the current tectonic environment.
In an anticline trap, tectonic uplift causes the reservoir's top surface to rise, with impermeable layers sealing the upper part and water bodies or impermeable rocks isolating the lower part.
This configuration creates ideal conditions for petroleum accumulation.




Despite its importance, the specialized and massive knowledge required to understand geologic map poses a high threshold.
Even geologists often struggle to quickly link and retrieve knowledge from external sources, such as geological, geographical, and seismological data.
This difficulty is amplified when dealing with private sources or large quantities of data, making efficient and systematic analysis of geologic map more challenging.
When considering leveraging AI for this, MLLMs have demonstrated strengths in general image understanding.
However, several challenges remain in geologic map understanding, which has the nature of cartographic generalization.
Specifically:
(1) \textbf{High Resolution}: it has extremely high resolutions, e.g., sometimes even reaching up to 10,000$^2$ pixels.
(2) \textbf{Multiple Associated Components}: it contains numerous components, many of which are interrelated.
(3) \textbf{Domain-specific Knowledge}: it consists of complex and symbolized geographical objects with diverse visual representations.
Additionally, a range of diverse AI capabilities is required, including detection, classification, segmentation, optical character recognition (OCR), cross-region understanding, and reasoning.
Consequently, the potential of MLLMs in understanding geologic map remains under-explored.


To address these challenges, we propose GeoMap-Bench, a benchmark for evaluating the performance of LLMs in geologic map understanding from 5 aspects: extracting, referring, grounding, reasoning, and analyzing.
Alongside GeoMap-Bench, we introduce GeoMap-Agent, an AI system specifically designed for geologic map understanding and analysis.
Its innovative design facilitates map digitization and enables question answering, extending even to downstream applications.
Specifically, GeoMap-Agent comprises three modules: Hierarchical Information Extraction (HIE), Domain Knowledge Injection (DKI), and Prompt-enhanced Question Answering (PEQA).
In summary, the main contributions are as follows:
\begin{itemize}
  \item We construct the GeoMap-Bench, which is the \textbf{first-ever} benchmark for evaluating the performance of MLLMs on geologic map understanding comprehensively.
  \item We propose the GeoMap-Agent, which is the \textbf{inaugural} AI agent for question answering of geologic map.
  Among it, several modules are designed for information structuring and deep thinking with domain-specific knowledge.
  \item The GeoMap-Agent achieves superior performance compared with the existing MLLMs, showcasing its potential to significantly enhance the efficiency and accuracy of geologic map understanding.
\end{itemize}


\section{Related Work}
\label{sec:related_work}


\noindent\textbf{Science Benchmark.}
GeoBench~\cite{deng2024k2} is specifically designed to test the geoscientific understanding of LLMs, focusing exclusively on text-based evaluations.
Charting New Territories benchmark~\cite{roberts2024charting} includes visual tasks aims at evaluating the geographic and geospatial capabilities of MLLMs.
OceanBench~\cite{bi2023oceangpt} is proposed to assess the capabilities of LLMs in performing tasks related to ocean science.
ScienceQA~\cite{lu2022learn}, a multimodal benchmark, consists of multiple-choice questions covering a wide range of science topics.
However, geology science is rarely considered in science QA.
LHRS-Bench~\cite{muhtar2024lhrs} is a comprehensive benchmark for thoroughly evaluating MLLMs in understanding remote sensing (RS) images.
Currently, despite the significance of geologic map, none of the existing benchmarks are designed to comprehensively understand them.


\noindent\textbf{Science Agent.}
GeoGPT~\cite{zhang2023geogpt} leverages Chain-of-Thought (COT)~\cite{wei2022chain} reasoning and a suite of GIS tools to tackle diverse geospatial tasks.
ChemCrow~\cite{bran2023chemcrow} is an LLM chemistry agent designed to accomplish tasks across organic synthesis, drug discovery, and materials design.
SocialSimulacra~\cite{park2022social} is a prototyping technique that generates a wide range of realistic social interactions that may emerge when a social computing system is populated.
DS-Agent~\cite{guo2024ds} is a novel automatic framework that harnesses a LLM agent and case-based reasoning (CBR). which can flexibly capitalize on expert knowledge from Kaggle~\cite{kaggle} and facilitate consistent performance improvement through the feedback mechanism.
Nevertheless, there is no work focusing on geologic map understanding.
Meanwhile, due to its challenges, other methods are difficult to apply effectively.

\noindent\textbf{Scientific LLM.}
Scientific models are being increasingly utilized to empower various scientific fields.
In geoscience domain,
GeoLLM~\cite{manvi2023geollm} enhances instruction prompt and is fine-tuned on GPT-3.5 turbo~\cite{ouyang2022training} to address geospatial prediction problems, such as population density prediction.
GeoGalactica~\cite{lin2023geogalactica} constructs a geological text corpus, and continues training on Galactica-30B~\cite{taylor2022galactica} for merely text-based geo-question answering.
K2~\cite{deng2024k2}, the first-ever LLM in geoscience, aims to align LLM responses to geoscience-related user queries, which only support text query.
In other scientific domains,
Clinical LLMs~\cite{singhal2023large} introduces a human evaluation framework and instruction prompt tuning.
MedGPT~\cite{kraljevic2021medgpt} applies Electronic Health Records (EHR) data and Named Entity Recognition tools to predict future medical events.
BioGPT~\cite{luo2022biogpt} is pre-trained on biomedical literature to facilitate biomedical text generation and mining.
OceanGPT~\cite{bi2023oceangpt} collects an extensive oceanic corpus and trains the first LLM specifically designed for ocean science tasks.
However, when considering applications in the geological field, there is a notable lack of related work focused on multimodal understanding, especially geologic map.


\noindent\textbf{General LLM.}
Large Language Models (LLMs)~\cite{brown2020language,yang2007paml,anthropic2024,achiam2023gpt,dubey2024llama,kaplan2020scaling} have rapidly developed recently, achieving significant breakthroughs across both general and specific domains.
To support cross-modality understanding, research has extended beyond pure text to include other modalities, such as CLIP~\cite{radford2021learning}, LLaVA~\cite{liu2024improved}, Qwen-VL~\cite{bai2023qwen}, and GPT-4V~\cite{achiam2023gpt} for image, as well as specific works like Gemini~\cite{team2023gemini} and Video-llava~\cite{lin2023video} in video, and AudioPaLM~\cite{rubenstein2023audiopalm} and VioLA~\cite{wang2023viola} in speech.
Among them, GPT-4o~\cite{openAI4o2024} stands out as one of the most powerful MLLMs.
Although, it's hard to be directly applied in geologic map understanding, we utilize it as the base model in GeoMap-Agent.

\section{GeoMap-Bench}
\label{sec:method_benchmark}


We construct a geologic map benchmark, GeoMap-Bench, to evaluate the performance of MLLMs on geologic map understanding across different abilities, the overview of it is as shown in Table~\ref{tab:geomap_benchmark}.
A detailed introduction is provided in the following subsections, and the evaluation metrics are defined in the Appendix.

\begin{table}[ht]
  \centering
  \begin{tabular}{@{}ll@{}}
    \toprule
    Property & Description \\
    \hline
    Source & \makecell[l]{USGS (English) \\ CGS (Chinese)} \\
    \hline
    Content & Image-question pair with annotated answer \\
    \hline
    Scale & 124 images and 3,864 questions \\
    \hline
    Resolution & \textbf{6,146$^2$} pixels on average \\
    \hline
    \makecell[l]{Question \\Type} & \makecell[l]{1. Multiple-choice question \\2. Fill-in-the-blank question \\3. Essay question} \\
    \hline
    \makecell[l]{Covering \\Ability} & \makecell[l]{1. Extracting \\2. Grounding \\3. Referring \\4. Reasoning \\5. Analyzing} \\
    \hline
    Defined Task & 25 tasks \\
    \bottomrule
  \end{tabular}
  \caption{The overview of GeoMap-Bench composition.}
  \label{tab:geomap_benchmark}
\vspace{-5px}
\end{table}


\subsection{Data Sources}
There are lots of reliable data sources for geologic maps, including United States Geological Survey (USGS), China Geological Survey (CGS), British Geological Survey (BGS), Geological Survey of Canada (GSC), European Geological Data Infrastructure (EGDI), OneGeology.
To construct a high-quality benchmark for geologic map, it is essential to utilize data that is \textbf{standardized} in format and \textbf{diverse} in content.
Specifically, the geologic maps from the target data sources should contain standardized components, as mentioned in the Appendix.
Additionally, the geologic maps should cover different physical regions with various geological features and be available in multiple languages.
Consequently, we select publicly available geologic maps from USGS and CGS as benchmark data sources.

\noindent\textbf{USGS Source.}
Geologic maps from the USGS source exhibit significant variation in drawing style and geologic components.
These maps are in rasterized format or ArcGIS~\cite{ormsby2004getting} format, with nearly 10,000 instances accessible.
They were published over a broad time span, from 1911 to 2024, and cover regions ranging from $\sim$67°W to $\sim$125°W longitude and $\sim$24°N to $\sim$49°N latitude.
Additionally, the scales of these maps range from 1:24,000 to 1:500,000.

\noindent\textbf{CGS Source.}
Geologic maps from the CGS source are standardized in drawing style and geological components.
These maps are in MapGIS~\cite{mapgis} format, with several thousands instances.
There are over 158 datasets, most of which were published after the 1980s.
They cover regions ranging from $\sim$72°E to $\sim$138°E longitude and $\sim$16°N to $\sim$56°N latitude, with scales mostly ranging from 1:50,000 to 1:250,000.


\subsection{Dataset Construction}
\label{sec:dataset_construction}

\begin{figure}
  \centering
  \includegraphics[width=3in]{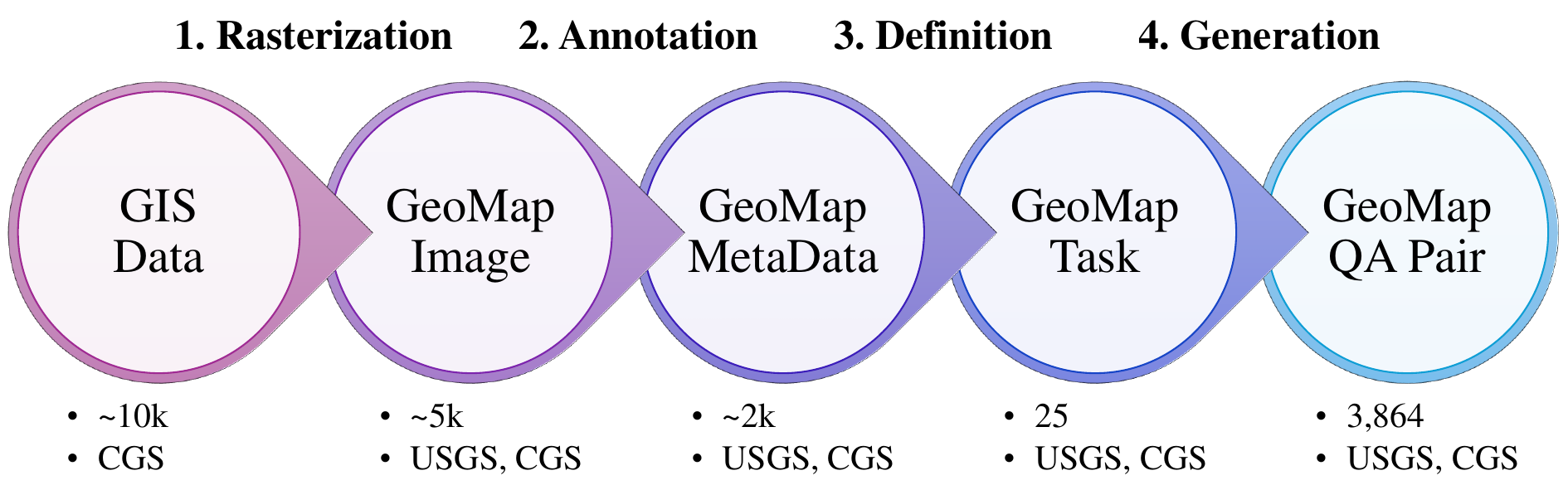}
  \hfill
  \caption{The pipeline of GeoMap-Bench construction.}
  \label{fig:benchmark_construction}
\vspace{-5px}
\end{figure}

\subsubsection{Rasterization}
Since the CGS map data is in the proprietary MapGIS~\cite{mapgis} format, unlike the ArcGIS format, we use MapGIS software~\cite{mapgis} to render them to rasterized images.
To expedite this process, a simulation program is developed that automates keyboard and mouse operations within the software.
We select the ``National 1:200,000 Digital Geologic Map (Public Version) Spatial Database" from the CGS maps for rasterization since their standard format in content, which includes 1,163 maps.
For the USGS maps, we choose those published after the 1990s in rasterized image format, totaling 6,415 maps.
After filtering for image quality, we obtain $\sim$5,000 rasterized images for further processing.


\subsubsection{Annotation}
After obtaining the rasterized maps, we manually annotate the metadata for each map, including the bounding box of each component, the basic information of each component (names, longitudes, scale, etc.), as well as the details of each legend unit (including bounding boxes, colors, texts, lithology, and stratigraphic age).
Additionally, we also record statistical information, such as the lithology composition, the area of each rock unit, and the presence of faults in different regions.
Following an accuracy cross-check of metadata, we have $\sim$2,000 images with corresponding metadata.


\subsubsection{Definition}
To thoroughly measure the performance of MLLMs on geologic map understanding, we collaborate with senior geologists to define measuring abilities across five aspects: extracting, grounding, referring, reasoning, and analyzing.
The details are as follows:

\begin{table*}[ht]
  \scriptsize
  \centering
  \begin{tabular}{@{}llll@{}}
    \toprule
    Ability & Task & Sampled Question & Type \\
    \hline
    \multirow{4}*{Extracting} & sheet name & What is the \textbf{name} of this map? & FITB \\
    & scale & What is the \textbf{scale} of this map? & FITB \\
    & lonlat & What is the \textbf{latitude and longitude} ranges of this map? & FITB \\
    & index map & What are the \textbf{neighboring areas} of this region? & FITB \\
    \hline
    \multirow{7}*{\makecell[l]{Grounding \\ (by name)}} & title by name & What is the location (bounding box) of the \textbf{title} component on the geologic map? & FITB \\
    & scale by name & What is the location (bounding box) of the \textbf{scale} component on the geologic map? & FITB \\
    & legend by name & What is the location (bounding box) of the \textbf{legend} component on the geologic map? & FITB \\
    & main map by name & What is the location (bounding box) of the \textbf{main map} component on the geologic map? & FITB \\
    & index map by name & What is the location (bounding box) of the \textbf{index map} component on the geologic map? & FITB \\
    & cross section by name & What is the location (bounding box) of the \textbf{cross section} component on the geologic map? & FITB \\
    & stratigraphic column by name & What is the location (bounding box) of the \textbf{stratigraphic column} component on the geologic map? & FITB \\
    \hdashline
    \multirow{14}*{\makecell[l]{Grounding \\ (by intention)}} & \makecell[l]{title by intention \\ $ $} & \makecell[l]{What location (bounding box) of the geologic map should I focus on \\ to \textbf{categorize, archive, and retrieve the geologic map}?} & FITB \\
    & \makecell[l]{scale by intention \\ $ $} & \makecell[l]{What location (bounding box) of the geologic map should I focus on \\ to \textbf{measure distances and understand the terrain scale}?} & FITB \\
    & \makecell[l]{legend by intention \\ $ $} & \makecell[l]{What location (bounding box) of the geologic map should I focus on \\ to \textbf{identify different geologic units and phenomena through the markings}?} & FITB \\
    & \makecell[l]{main map by intention \\ $ $} & \makecell[l]{What location (bounding box) of the geologic map should I focus on \\ to \textbf{identify the distribution of specific geologic resources}?} & FITB \\
    & \makecell[l]{index map by intention \\ $ $} & \makecell[l]{What location (bounding box) of the geologic map should I focus on \\ to \textbf{identify the names of adjacent geologic map sheets in different directions}?} & FITB \\
    & \makecell[l]{cross section by intention \\ $ $} & \makecell[l]{What location (bounding box) of the geologic map should I focus on \\ to \textbf{understand geologic structures from a three-dimensional perspective}?} & FITB \\
    & \makecell[l]{stratigraphic column by intention \\ $ $} & \makecell[l]{What location (bounding box) of the geologic map should I focus on \\ to \textbf{understand the deposition or formation time of different strata to help determine their age}?} & FITB \\
    \hline
    \multirow{2}*{Referring} & color by rock & In this geologic map, what \textbf{legend color} is used to represent the 'Quartz-sericite phyllite and schist' rock name? & MCQ \\
    & rock by color & In this geologic map, what is the \textbf{rock name} whose legend color is closest to \textbf{\textcolor[HTML]{5D1C1C}{\#5D1C1C}}? & MCQ \\
    \hline
    \multirow{4}*{Reasoning} & area comparison & Regarding the \textbf{rock name} in main map, which one \textbf{ranks third} in area among 4 choices? & MCQ \\
    & \makecell[l]{fault existence \\ $ $} & \makecell[l]{If the area represented by the geologic map is equally divided into a 3x3 grid, \\ is there a \textbf{fault} in the grid located in the \textbf{Northeast} direction?} & MCQ \\
    & lithology composition & In this geologic map, which type of \textbf{primary lithology} has the \textbf{largest proportion}? & MCQ \\
    & lonlat localization & Can you infer the most likely \textbf{title} of the map in which \textbf{(longitude:-81.5, latitude:35.25)} is located? & MCQ \\
    \hline
    \multirow{1}*{Analyzing} & earthquake risk &  Based on this geologic map, please analyze the \textbf{seismic risk} in this area from possibility and societal impact? & EQ \\
    \bottomrule
  \end{tabular}
  \caption{Defined tasks in GeoMap-Bench.
  The question format is randomly selected from a format pool for each task.
  For example, both ``What's the scale of this map?" and ``Can you provide the scale of this map?" are variations of the same task of extracting the map scale.
  Meanwhile, each component corresponds to an intention pool, from which an intention is randomly chosen per question in a grounding-by-intention task.
  The question types ``FITB," ``MCQ," and ``EQ" represent fill-in-the-blank, multiple-choice, and essay questions, respectively.
  }
  \label{tab:geomap_benchmark}
  \vspace{-10px}
\end{table*}

\noindent\textbf{Extracting.}
It evaluates the ability to accurately extract basic information from the map, such as the title, scale, and latitude coordinates.

\noindent\textbf{Grounding.}
It measures the capability to precisely locate components on the map based on their names or intentions.

\noindent\textbf{Referring.}
It assesses the skill in linking names to their corresponding properties, such as identifying the rock name by its legend color.

\noindent\textbf{Reasoning.}
It evaluates the ability to perform high-level logical tasks that require connecting information across components or incorporating external knowledge.

\noindent\textbf{Analyzing.}
It assesses the capability to comprehensively interpret a given topic on the map and provide detailed and meaningful insights from various perspectives.

To concretize these abilities, we delineate specific tasks for each aspect, totaling 25 tasks, which is defined in Table~\ref{tab:geomap_benchmark}.
After this process, we compile a dataset comprising $\sim$5,000 questions, ensuring the completeness of each map and a balanced task distribution of questions.


\begin{figure}
  \centering
     \includegraphics[width=3.3in]{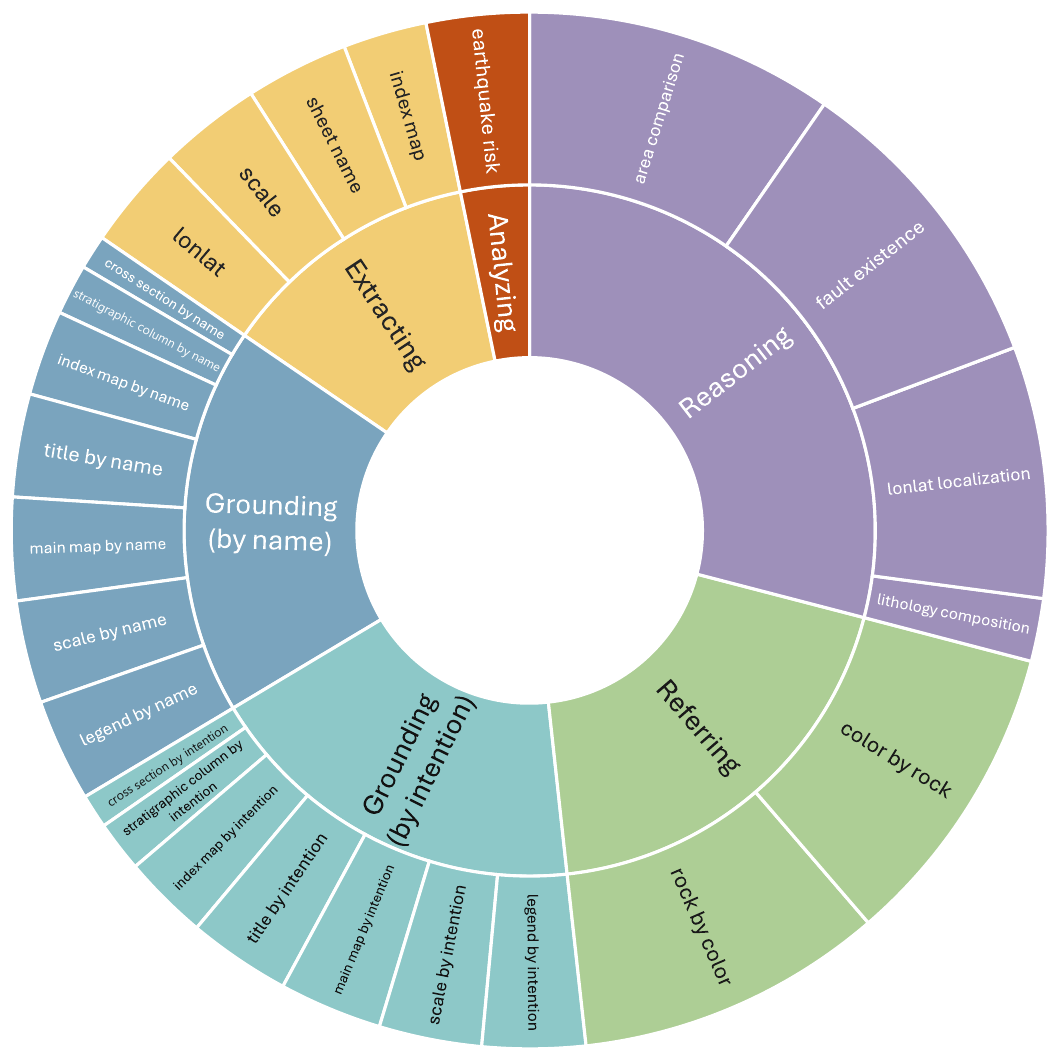}
  \hfill
\vspace{-10px}
  \caption{The distribution of questions in GeoMap-Bench.
  It consists of 25 task types, which measure MLLMs abilities across 5 aspects.
  The questions within these different task types are relatively evenly distributed, as indicated by the area of each task.
  }
  \label{fig:benchmark_distribution}
\vspace{-25px}
\end{figure}

\subsubsection{Generation}
Based on the annotated metadata of geologic maps, we generate the ground-truth answer for each question through retrieving, calculating, and statistics.
To ensure the quality of ground-truth answers, we enlist senior geologists to review and correct all the map-question pairs.
Eventually, this process results in a dataset comprising 124 maps and 3,864 questions with ground-truth answers.
The distribution of questions in GeoMap-Bench across different abilities and tasks is illustrated in Figure~\ref{fig:benchmark_distribution}.



\begin{figure*}
  \centering
     \includegraphics[width=6.8in]{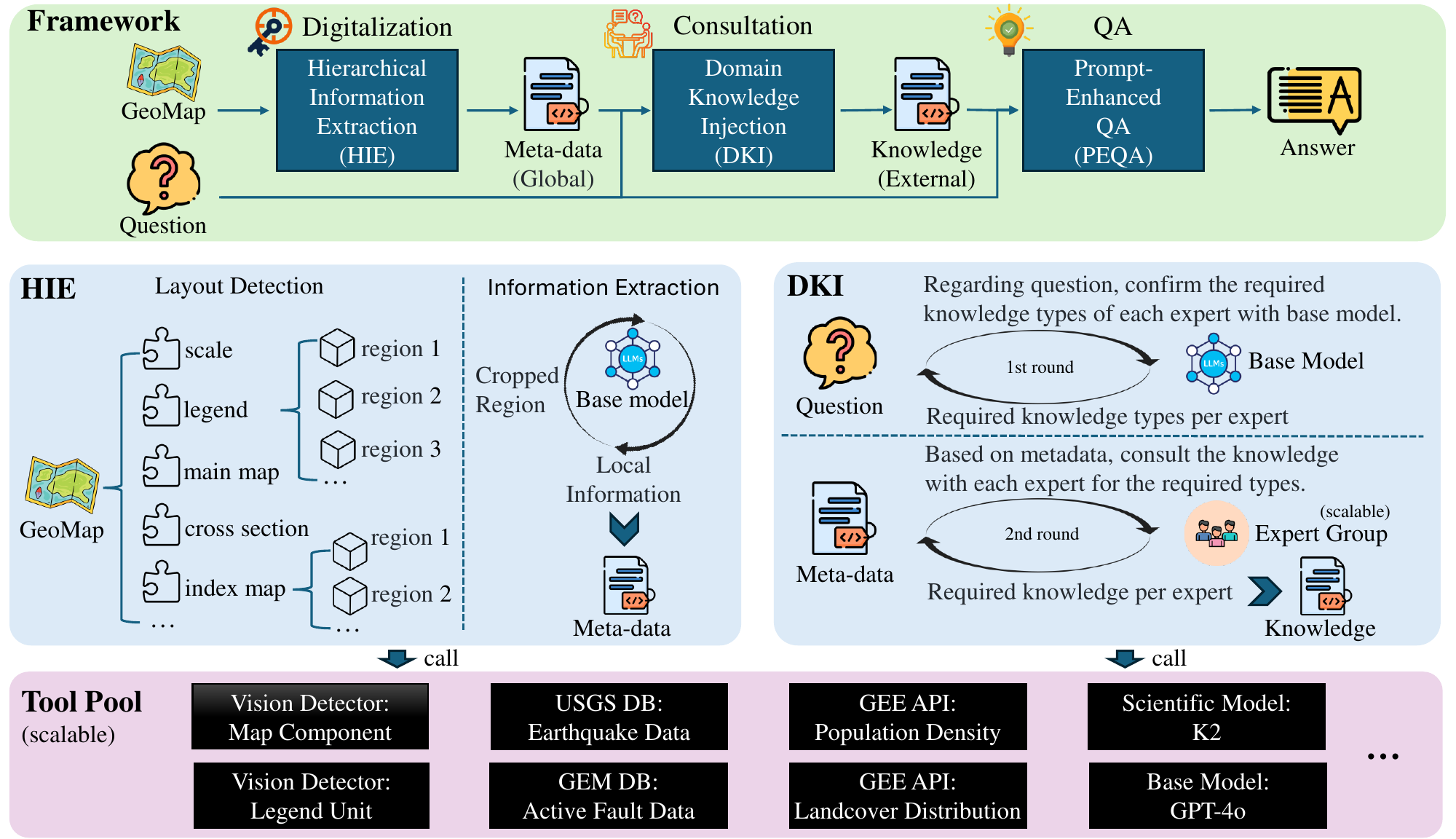}
  \hfill
  \caption{The framework of GeoMap-Agent.
  It consists of three modules: Hierarchical Information Extraction (HIE), Domain Knowledge Injection (DKI), and Prompt-enhanced Question Answering (PEQA).
  HIE digitizes the map by extracting the global information hierarchically.
  DKI involves expert group as consultants to provide required domain-specific knowledge for each question.
  PEQA enhances the QA prompt using the extracted metadata and injected knowledge, as detailed in Section~\ref{sec:module_PEQA}.
  All the modules are empowered by a tool pool.
  Both the expert group and tool pool are scalable.
  }
  \label{fig:GeoMap_Agent}
\vspace{-10px}
\end{figure*}


\section{GeoMap-Agent}
\label{sec:method_agent}


Scientific diagrams in certain scenarios often exhibit attributes such as high resolution, multiple associated components, and the need for domain knowledge.
These attributes pose significant challenges for image understanding of MLLMs.
For instance, geologic map is poorly understood even by the powerful GPT-4o, as demonstrated in Table~\ref{tab:main_experiment}.
To tackle these challenges, we propose an MLLM-based paradigm for image understanding.
GeoMap-Agent, illustrated in Figure~\ref{fig:GeoMap_Agent}, serves as an example of this paradigm.


\begin{figure}[htbp]
  \centering
  \includegraphics[width=3.3in]{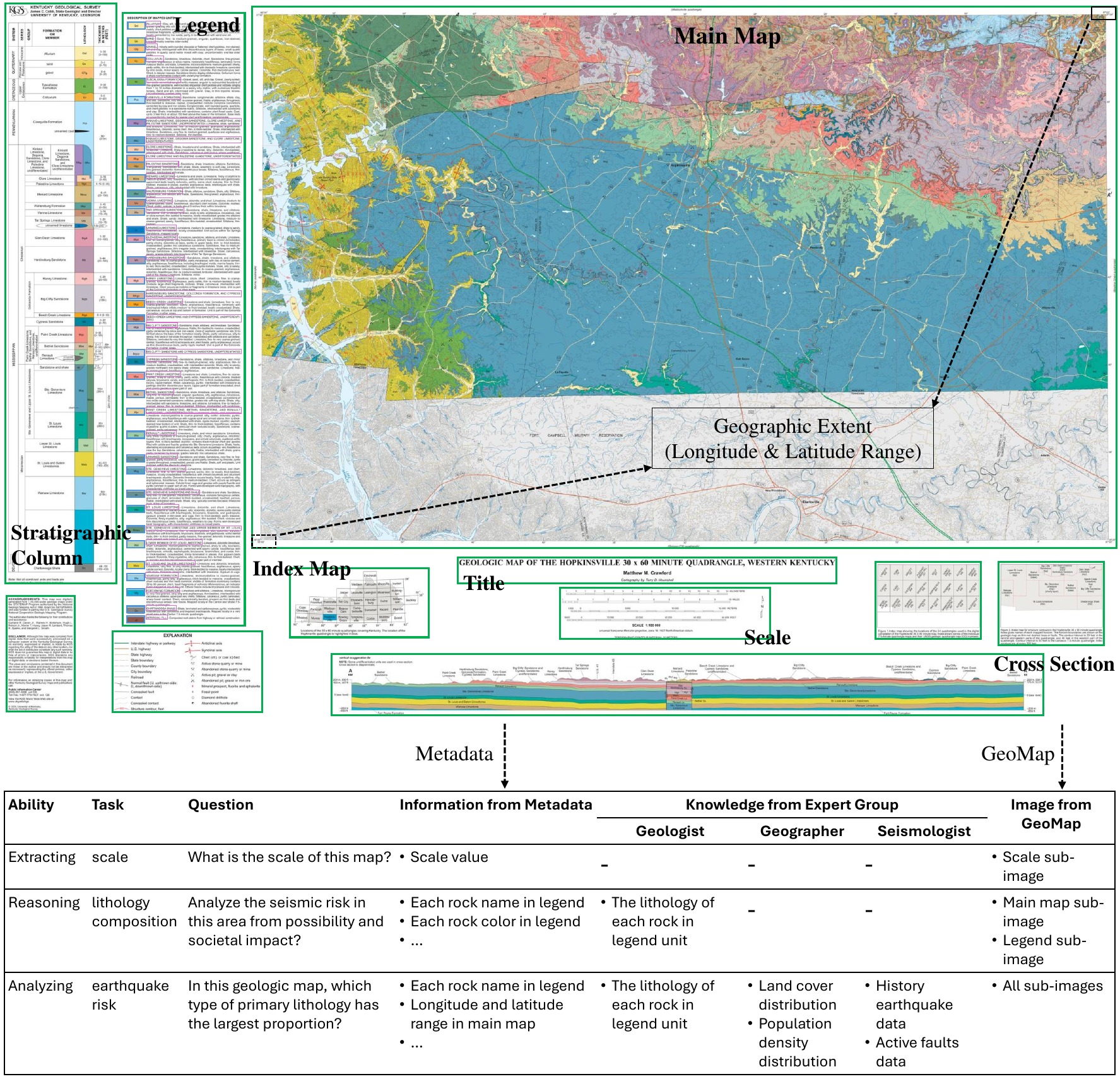}
  \hfill
\vspace{-8px}
  \caption{Example of geologic map and its question answering.
  The geologic map contains numerous semantic components with specific information.
  Through HIE, the vision detectors identify these regions, and the base model extracts information region by region, which is then merged into global metadata.
  In the DKI module, each question acquires specific knowledge by consulting the expert group.
  Finally, the PEQA module utilizes metadata and knowledge to perform prompt-enhanced QA.
  }
  \label{fig:geomap}
\vspace{-8px}
\end{figure}

\subsection{Modules}
\label{sec:method_agent_modules}
The framework of GeoMap-Agent consits of three modules, Hierarchical Information Extraction (HIE), Domain Knowledge Injection (DKI), and Prompted-enhanced Question Answering (PEQA).


\subsubsection{HIE}
\label{sec:module_HIE}
Extracting information from a high-resolution image often leads to poor performance for MLLMs.
The Hierarchical Information Extraction (HIE) module mitigates this by employing a ``divide and conquer" strategy.
In the ``divide" stage, the entire image is treated as the root of a tree, and it is hierarchically divided into sub-images.
To ensure each sub-image represents a semantically independent component, a map component detector and legend unit detector from the tool pool are progressively applied.
During the ``conquer" stage, a base model, currently GPT-4o, is applied to extract local information from each sub-image.
This model can be replaced with a more powerful one as it becomes available.
To ensure comprehensive information extraction, the K2~\cite{deng2024k2}, geology specialist model, from the tool pool is utilized to provide a detailed information list for each component, enhancing the extraction prompt.
Finally, the extracted information from each sub-image is aggregated into the global metadata.
This process effectively forms a structured representation of geologic maps.


\subsubsection{DKI}
\label{sec:module_DKI}
The Domain Knowledge Injection (DKI) module supplies the essential field knowledge for question answering, particularly for questions that require reasoning and analyzing.
The DKI module operates in two steps:
First, for the given question, the base model is confirmed with to determine whether specific types of knowledge from each expert in the group are needed.
Second, for the required knowledge types, the relevant experts are consulted one by one to acquire the corresponding knowledge, some of which is linked through the latitude and longitude range extracted in HIE.
Currently, our expert group includes a geologist, a geographer, and a seismologist, each of whom can leverage related tools from the tool pool.
The expert group can be easily extended based on specific requirements.


\subsubsection{PEQA}
\label{sec:module_PEQA}
In the previous two modules, the global metadata (internal information) of geologic map and the required knowledge (external knowledge) specific to the question about this map are obtained.
Based on them, the prompt-enhanced question answering (PEQA) module enhances the prompt from different aspects: context, chain of thought (CoT)~\cite{wei2022chain}, few-shot learning, and attention-like design.
(1). We provide the digitized metadata and required knowledge as context in the prompt.
(2). We require the base model to respond with not only answer but also reasoning, which encouraging deeper thinking.
(3). We supply an example answer in the prompt and instruct the base model to respond in JSON format to ensure proper formatting.
(4). We crop the relevant components of geologic map to the given question and include the cropped images in the prompt.
By leveraging these prompt designs, the performance is further improved.


\subsection{Expert Group}
Inspired by the working models of human scientists, interdisciplinary group collaboration is employed.
The expert group comprises AI experts from various scientific fields, which provides domain-specific knowledge for GeoMap-Agent, as illustrated in Figure~\ref{fig:GeoMap_Agent}.
Each expert specializes in a distinct field mastering unique types of knowledge.
Both the knowledge types of each expert and the composition of the expert group are scalable.

\noindent\textbf{Geologist.}
The geologist specializes in geologic map knowledge, including the composition of geologic map, the information schema of each geologic component, the table of stratigraphic age, and the table of lithology.

\noindent\textbf{Geographer.}
The geographer focuses on distributions of land cover and population density covering global regions.

\noindent\textbf{Seismologist.}
The seismologist concentrates on both history earthquake data and active faults data worldwide.


\subsection{Tool Pool}
The tool pool provides various functionalities for modules of GeoMap-Agent.
It currently contains more than 8 tools and is scalable to accommodate additional tools as needed.

\noindent\textbf{Vision Detectors.}
Two vision detectors~\cite{wang2024yolov10} are developed that assist in analyzing the layout of geologic map.
The map component detector identifies 7 components as outlined in Appendix, while the legend unit detector identifies all text units and color units within the legend component, aiding in rock name recognition and rock color detection.

\noindent\textbf{GEE APIs.}
Two APIs in Google Earth Engine (GEE)~\cite{gorelick2017google} are included to retrieve the population density~\cite{sorichetta2015high,gaughan2013high} and land cover~\cite{zanaga2022esa} distributions of an area based on the specified latitude and longitude range in geologic map.

\noindent\textbf{Scientific DBs.}
Two scientific databases are integrated: one for accessing historical earthquake records from the USGS database~\cite{earthquake1977USGS} and the other for active fault records from the GEM database~\cite{styron2020gem}.
Both of them can be retrieved based on the specified latitude and longitude coordinates.

\noindent\textbf{AI Models.}
Two models are involved:
the K2~\cite{deng2024k2} scientific model offers geologic-specific knowledge, particularly concerning geologic map;
the GPT-4o~\cite{openAI4o2024} base model is utilized for information extraction, question answering, etc.

\noindent\textbf{Others.}
Other tools are described in the Appendix.


\section{Experiment}
\label{sec:experiment}


\subsection{Performance on Benchmark}



\begin{table*}[ht]
  \centering
  \begin{tabular}{@{}llcccccc@{}}
    \toprule
    Dataset & Method & Extracting & Grounding & Referring & Reasoning & Analyzing & Overall \\
    \hline
    \multirow{9}*{\makecell{USGS \\ Set}} & Random & 0 & 0 & 0.250 & 0.250 & 0 & 0.100 \\
    & QWen-chat~\cite{bai2023qwen} & 0.050 & 0.003 & 0.253 & 0.442 & 0.250 & 0.199 \\
    & GLM-4v-9b~\cite{glm2024chatglm} & 0.050 & 0.010 & 0.258 & 0.212 & 0.600 & 0.226 \\
    & Idefics-9b-instruct~\cite{laurenccon2024obelics} & 0.025 & 0.000 & 0.247 & 0.260 & 0.333 & 0.173 \\
    & Cogvlm2-llama3-chat-19B~\cite{hong2024cogvlm2} & 0.033 & 0.000 & 0.189 & 0.177 & 0.067 & 0.093 \\
    & Monkey-chat~\cite{li2024monkey} & 0.042 & 0.010 & 0.213 & 0.349 & 0.267 & 0.176 \\
    & GPT-4o-mini~\cite{openAI4o2024} & 0.183 & 0.050 & 0.278 & 0.456 & 0.512 & 0.295 \\
    & GPT-4o~\cite{openAI4o2024} & 0.208 & 0.100 & 0.398 & 0.494 & 0.683 & 0.376 \\
    & \textbf{GeoMap-Agent (Ours)} & \textbf{0.887} & \textbf{0.935} & \textbf{0.949} & \textbf{0.581} & \textbf{0.817} & \textbf{0.833} \\
    \hline
    \multirow{9}*{\makecell{CGS \\ Set}} & Random & 0 & 0 & 0.250 & 0.250 & 0 & 0.100 \\
    & QWen-chat~\cite{bai2023qwen} & 0.000 & 0.003 & 0.264 & 0.334 & 0.457 & 0.211 \\
    & GLM-4v-9b~\cite{glm2024chatglm} & 0.295 & 0.076 & 0.234 & 0.366 & 0.468 & 0.287 \\
    & Idefics-9b-instruct~\cite{laurenccon2024obelics} & 0.000 & 0.000 & 0.235 & 0.134 & 0.457 & 0.165 \\
    & Cogvlm2-llama3-chat-19B~\cite{hong2024cogvlm2} & 0.205 & 0.000 & 0.236 & 0.156 & 0.415 & 0.202 \\
    & Monkey-chat~\cite{li2024monkey} & 0.031 & 0.002 & 0.248 & 0.145 & 0.457 & 0.176 \\
    & GPT-4o-mini~\cite{openAI4o2024} & 0.204 & 0.102 & 0.287 & 0.474 & 0.491 & 0.311 \\
    & GPT-4o~\cite{openAI4o2024} & 0.230 & 0.157 & 0.359 & 0.521 & 0.542 & 0.361 \\
    & \textbf{GeoMap-Agent (Ours)} & \textbf{0.777} & \textbf{0.906} & \textbf{0.824} & \textbf{0.595} & \textbf{0.846} & \textbf{0.789} \\
    \hline
    \multirow{2}*{\makecell{All \\ Sets}} & Random & 0 & 0 & 0.250 & 0.250 & 0 & 0.100 \\
    & GPT-4o & 0.219 & 0.128 & 0.378 & 0.507 & 0.612 & 0.369 \\
    & \textbf{GeoMap-Agent} & \textbf{0.832} & \textbf{0.920} & \textbf{0.886} & \textbf{0.588} & \textbf{0.831} & \textbf{0.811} \\
    \bottomrule
  \end{tabular}
  \caption{Evaluation of different methods on GeoMap-Bench.
  The publicly available MLLMs perform poorly across all metrics.
  In contrast, our GeoMap-Agent demonstrates superior performance on both subsets, significantly surpassing all other methods in every aspect.
  }
  \label{tab:main_experiment}
\vspace{-10px}
\end{table*}

We conducted comparison experiments on GeoMap-Bench using various methods, including the publicly available API, like GPT-4o~\cite{openAI4o2024}, and open-source models, such as QWen-chat~\cite{bai2023qwen}.
As shown in Table~\ref{tab:main_experiment}, GeoMap-Agent consistently achieves the best performance across different subsets (USGS and CGS) on all aspects of ability, including extracting, grounding, referring, reasoning, and analyzing.
Specifically, in our designed benchmark, public MLLMs exhibit significant weaknesses, particularly in extracting, grounding, and reasoning.
Although GPT-4o performs the best among public MLLMs, its grounding ability remains subpar, resulting in an overall benchmark score of less than 0.5.
In contrast, GeoMap-Agent demonstrates outstanding performance in basic abilities, such as 0.832 in extracting, 0.920 in grounding, and 0.886 in referring.
It also shows relative high performance in advanced abilities, like 0.588 in reasoning, and 0.831 in analyzing.
In summary, GeoMap-Bench comprehensively evaluates different methods for geologic map understanding, and GeoMap-Agent enables a thorough understanding of geologic map, as verified by its performance.


\subsection{Ablation Study}
To thoroughly analyze the GeoMap-Agent, more experiments are conducted for different purposes, such as the contribution of each module, the adaptability to other base models, and the performance under lower resolutions.
Additional experiments can be found in the Appendix.


\subsubsection{Contribution on Different Modules}
General LLMs encounter several challenges in understanding geologic map,
and the three modules proposed in Section~\ref{sec:method_agent_modules} mitigate these challenges.
To verify their effectiveness, we conduct experiments to assess the contribution of each module by applying them partly.
As shown in Table~\ref{tab:ablation_module}, each module enhances the abilities of GeoMap-Agent from different perspectives.
For example, HIE significantly improves basic abilities by addressing challenges related to high resolution and multiple associated components, DKI further enhances advanced abilities by incorporating domain knowledge, while PEQA boosts overall performance.
GeoMap-Agent achieves optimal results when all these modules are applied.

\begin{table}[ht]
  \centering
  \begin{tabular}{@{}llccccc@{}}
    \toprule
    Dataset & Ability & \makecell{HIE,\\DKI,\\PEQA} & \makecell{HIE,\\DKI} & HIE & None \\
    \hline
    \multirow{6}*{\makecell{USGS \\ Set}} & Ext. & 0.887 & 0.745 & 0.741 & 0.208 \\
    & Gro. & 0.935 & 0.760 & 0.747 & 0.100 \\
    & Ref. & 0.949 & 0.835 & 0.818 & 0.398 \\
    & Rea. & 0.581 & 0.573 & 0.526 & 0.494 \\
    & Ana. & 0.817 & 0.786 & 0.691 & 0.683 \\
    & Ove. & 0.833 & 0.739 & 0.704 & 0.376 \\
    \hline
    \multirow{6}*{\makecell{CGS \\ Set}} & Ext. & 0.777 & 0.627 & 0.596 & 0.230 \\
    & Gro. & 0.906 & 0.810 & 0.796 & 0.157 \\
    & Ref. & 0.824 & 0.755 & 0.755 & 0.359 \\
    & Rea. & 0.595 & 0.596 & 0.550 & 0.521 \\
    & Ana. & 0.846 & 0.783 & 0.572 & 0.542 \\
    & Ove. & 0.789 & 0.714 & 0.653 & 0.361 \\
    \bottomrule
  \end{tabular}
  \caption{Contributions of each module in GeoMap-Agent on GeoMap-Bench.
  All other settings remain the same, including the use of GPT-4o as base model.
  The symbols of ``Ext.", ``Gro.", ``Ref.", ``Rea.", ``Ana.", and ``Ove." represent extracting, grounding, referring, reasoning, analyzing, and overall respectively in this and following tables.}
  \label{tab:ablation_module}
\end{table}


\subsubsection{Adaptability to Different Base Models}
According to the GeoMap-Agent framework, the base model can theoretically be any MLLM.
To verify the adaptability of different base models, we test the GeoMap-Agent framework using GPT-4o~\cite{openAI4o2024} or GPT-4o mini~\cite{openAI4o2024}.
The results, shown in Table~\ref{tab:ablation_base}, indicate varying degrees of improvement by different base model.
Inferred from this, the base model is not limited to GPT-4o~\cite{openAI4o2024} and can be upgraded when more powerful MLLMs become available.

\begin{table}[ht]
  \centering
  \begin{tabular}{@{}llccc@{}}
    \toprule
    Dataset & Ability & GPT-4o~\cite{openAI4o2024} & GPT-4o-mini~\cite{openAI4o2024} \\
    \hline
    \multirow{6}*{\makecell{USGS \\ Set}} & Ext. & 0.887 & 0.737 \\
    & Gro. & 0.935 & 0.841 \\
    & Ref. & 0.949 & 0.813 \\
    & Rea. & 0.581 & 0.522 \\
    & Ana. & 0.817 & 0.791 \\
    & Ove. & 0.833 & 0.740 \\
    \hline
    \multirow{6}*{\makecell{CGS \\ Set}} & Ext. & 0.777 & 0.561 \\
    & Gro. & 0.906 & 0.735 \\
    & Ref. & 0.824 & 0.637 \\
    & Rea. & 0.595 & 0.539 \\
    & Ana. & 0.846 & 0.825 \\
    & Ove. & 0.789 & 0.659 \\
    \bottomrule
  \end{tabular}
  \caption{Performance comparison of different base model in GeoMap-Agent on GeoMap-Bench.
  All other settings remain the same, including the use of all 3 modules.}
  \label{tab:ablation_base}
\vspace{-10px}
\end{table}


\subsubsection{Performance across Different Resolutions}
High resolution presents a significant challenge for MLLMs in understanding geologic map.
To tackle this, the HIE in GeoMap-Agent crops the semantic regions of map into sub-images for information extraction.
As illustrated in Table~\ref{tab:ablation_resolution}, reducing the resolution of geologic map does not improve performance.
Therefore, we conclude that the improvement provided by HIE is not due to the direct reduction in resolution, but rather from the “divide and conquer” strategy.

\begin{table}[ht]
  \centering
  \begin{tabular}{@{}llccc@{}}
    \toprule
    \multirow{2}*{Dataset} & \multirow{2}*{Ability} & \multicolumn{3}{c}{Resolution Scale} \\
    \cmidrule(lr){3-5}
    & & 1 & 1/2 & 1/4 \\
    \hline
    \multirow{6}*{\makecell{USGS \\ Set}} & Ext. & 0.208 & 0.183 & 0.183 \\
    & Gro. & 0.100 & 0.083 & 0.083 \\
    & Ref. & 0.398 & 0.415 & 0.393 \\
    & Rea. & 0.494 & 0.479 & 0.468 \\
    & Ana. & 0.683 & 0.350 & 0.383 \\
    & Ove. & 0.376 & 0.302 & 0.302 \\
    \hline
    \multirow{6}*{\makecell{CGS \\ Set}} & Ext. & 0.230 & 0.230 & 0.220 \\
    & Gro. & 0.157 & 0.176 & 0.165\\
    & Ref. & 0.359 & 0.355 & 0.366 \\
    & Rea. & 0.521 & 0.519 & 0.513 \\
    & Ana. & 0.542 & 0.534 & 0.527 \\
    & Ove. & 0.361 & 0.362 & 0.358 \\
    \bottomrule
  \end{tabular}
  \caption{Performance comparison under different resolution scales.
  All other settings remain the same, including the use of GPT-4o as base model and the application of all 3 modules.
  Resolution scale 1/2 indicates resizing each original geologic map to 1/2 its original size.
  }
  \label{tab:ablation_resolution}
\vspace{-15px}
\end{table}


\section{Discussion}
\label{sec:discussion}


\noindent\textbf{GeoMap-Bench.}
Although GeoMap-Bench comprehensively evaluates geologic map understanding from various aspects, there is room for extending more abilities and tasks, especially those requiring analysis and external domain-specific knowledge, such as natural resource exploration.
Additionally, for essay-type questions, there is potential to improve evaluation methods to enhance both effectiveness and efficiency from different levels.


\noindent\textbf{GeoMap-Agent.}
While GeoMap-Agent performs well in GeoMap-Bench, several limitations are evident based on the framework and specific task evaluations:
(1) Despite incorporating domain knowledge from the expert group, conducting reasoning for question answering, such as fault detection and lithology composition, remains challenging.
(2) It struggles to recognize rocks with complex patterns or similar colors in the legend, especially in CGS subset.
To further improve its performance, we could either extend the expert group and tool pool or explore supervised fine-tuning on MLLMs.
Moreover, the framework of our agent could be treated as a paradigm for scenarios facing similar challenges, such as high resolution, multiple associated components, and the need for domain-specific knowledge.


\section{Conclusion}
\label{sec:conclusion}

Geologic map is a crucial scientific diagram in geology field with numerous applications.
This work leverages MLLMs to investigate and promote the performance of its understanding.
To elaborate,
the GeoMap-Bench aims to quantify the ability of geologic map understanding across different aspects,
and the GeoMap-Agent is a comprehensive framework designed to understand geologic map by question answering, such as lithology composition and earthquake risk assessment.
The experiments demonstrate that GeoMap-Agent significantly overperforms publicly available MLLMs on GeoMap-Bench.
These findings highlight that GeoMap-Agent effectively addresses the challenges faced by MLLMs, such as handling high resolution, multiple components with intricate interrelations, and the need for domain knowledge.
On the one hand,
it is expected to help geologists efficiently and comprehensively analyze the vast number of geologic maps worldwide.
On the other hand,
we believe that the GeoMap-Agent paradigm could also be applied to scenarios with similar challenges.


\clearpage
\setcounter{table}{0}   
\setcounter{figure}{0}
\setcounter{section}{0}
\setcounter{equation}{0}

\renewcommand{\thesection}{A\arabic{section}}
\renewcommand{\thetable}{A\arabic{table}}
\renewcommand{\thefigure}{A\arabic{figure}}
\renewcommand{\thesection}{A\arabic{section}}
\renewcommand{\theequation}{A\arabic{equation}}
\maketitlesupplementary
\appendix



\section{Geologic Map}
\label{sec:map_intro}
\begin{figure}[htbp]
  \centering
  \includegraphics[width=3.3in]{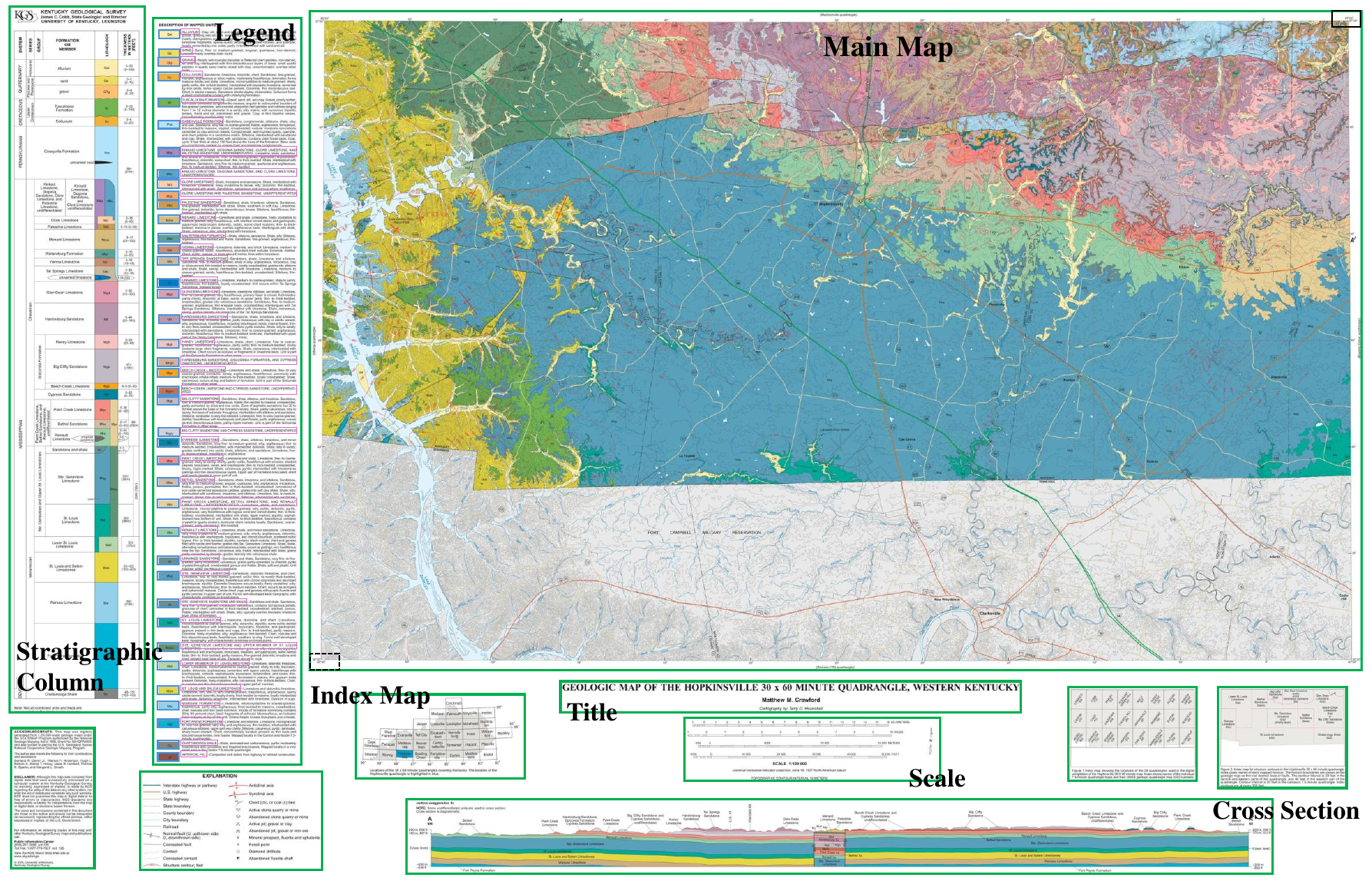}
  \hfill
  \caption{Example of a geologic map and its components.
  All components and legend units are enclosed within bounding boxes for interpretive understanding.
  }
  \label{fig:geomap_comps}
\end{figure}

Geologic map is a specialized type of map that depicts the distribution, characteristics, and chronological relationships of rock units as well as the occurrence of structural features such as faults and folds.
These maps are essential tools for geologists and earth scientists as they provide a visual representation of the geological characteristics of a specific area.
Typically, as shown in Figure~\ref{fig:geomap_comps}, a geologic map comprises several key elements, including the title, scale, legend, main map, index map, cross section, stratigraphic column, and other components.
These elements collectively contribute to the coherence and utility of a geologic map.
Specifically, please refer to the following content.

\noindent\textbf{Title} indicates the physical region, map type, author, and other pertinent information.

\noindent\textbf{Scale} demonstrates the relationship between distances on the map and physical distances on the ground.

\noindent\textbf{Legend} explains the symbols and colors used to represent different rock types, ages, and geological features.
For detailed information on the legend units, refer to the legend component in Figure~\ref{fig:geomap_comps}.

\noindent\textbf{Main Map} depicts the geological characteristics of the mapped area, including distributions of rock types, ages, folds, and faults.

\noindent\textbf{Index Map} illustrates the relationship and connection to neighboring regions.

\noindent\textbf{Cross Section} provides a vertical slice through the Earth, showing the arrangement of rock units below the surface.

\noindent\textbf{Stratigraphic Column} displays the sequence, thickness, and types of rock layers present in a particular area.

\noindent\textbf{Other Components} Besides the above 7 key components frequently found in geologic maps, there are additional supplementary components that provide further geological explanation for the region.

\section{Evaluation Metrics}
\label{sec:eval_metrics}
The metrics are designed to measure the quality of answers generated by AI-based methods for each question in GeoMap-Bench.

\subsection{Overall Score}
\label{sec:overall_score}
$S_{all}$ is the overall score of an AI-based method on GeoMap-Bench, where $M$ denotes the number of abilities to be measured in it, including extracting, grounding, referring, reasoning, and analyzing.

\begin{equation}
S_{all} = \frac{1}{M} \sum_{i=1}^{M}S_{i}(T, Q, A, L)
\label{equation:score_overall}
\end{equation}

\subsection{Ability Score}
\label{sec:ability_score}
$S_{i}$ is the ability score of an AI-based method measured for $i$-th ability in GeoMap-Bench,
where $N$ represents the number of questions pertaining to that ability.
$T$, $Q$, $A$, and $L$ indicate the sets of question types, questions, AI-responded answers, and expert-labeled answers respectively.
The $j$-th instance of these sets are denoted as $t_j$, $q_j$, $a_j$, and $l_j$.

\begin{equation}
S_{i}(T, Q, A, L) =
\frac{1}{N} \sum_{j=1}^{N}
S_{i, t_j}(q_j, a_j, l_j)
\label{equation:score_ability}
\end{equation}

\subsection{Type Score}
\label{sec:type_score}
$S_{i, t_j}$ is the type score for the $j$-th question type within the $i$-th ability.
This score can correspond to one of the following types: $S_{mcq}$ for multiple-choice questions, $S_{fitb}$ for fill-in-the-blank questions, and $S_{eq}$ for essay questions.

\noindent\textbf{Multiple-choice Question.}
$S_{mcq}$ is the type score of a multiple-choice question, where $q$, $a$, $l$ are a element of sets $Q$, $A$, $L$ respectively.

\begin{equation}
S_{mcq}(q, a, l) =
\begin{cases} 
1.0, & \text{$a$ = $l$} \\
0.0, & \text{otherwise}
\end{cases}
\label{equation:score_mcq}
\end{equation}

\noindent\textbf{Fill-in-the-blank Question.}
$S_{fitb}$ is the type score of a fill-in-the-blank question.

\begin{equation}
S_{fitb}(q, a, l) =
\begin{cases}
IoU_{det}(a, l), & \small\text{all grounding tasks} \\
IoU_{set}(a, l), & \small\text{set extracting tasks} \\
S_{mcq}(q, a, l), & \small\text{otherwise}
\end{cases}
\label{equation:score_fitb}
\end{equation}

where all grounding tasks encompass tasks of both grounding by name and grounding by intention, and set extracting tasks include tasks of index map extracting and longitude-latitude extracting.

$IoU_{det}$ is the intersection over union metric to evaluate the accuracy of a predicted bounding box against the ground-truth bounding box.

\begin{equation}
IoU_{det}(b_1, b_2) = \frac{I(b_1, b_2)}{U(b_1, b_2)}
\label{equation:iou_det}
\end{equation}

where $b_1$ and $b_2$ are two bounding boxes.
$I$ and $U$ are functions to calculate the intersection area and union area of two bounding boxes respectively.

$IoU_{set}$ is the intersection over union metric to evaluate the overlap of two sets.

\begin{equation}
IoU_{set}(A, B) = \frac{|A \cap B|}{|A \cup B|}
\label{equation:iou_set}
\end{equation}

where $A$ and $B$ are two sets, which could be either discrete, such as neighboring regions, or continuous, like longitude and latitude range.

\noindent\textbf{Essay Question.}
$S_{eq}$ is the type score of an essay question.
To avoid any order-related bias of answer-judging agent, it is measured twice per question by keeping and switching the order of two answers.

\begin{equation}
S_{eq}(q, a, l) = \frac{1}{2} (1 - J(q, l, a) + J(q, a, l))
\label{equation:score_eq}
\end{equation}

$J$ is a answer-judging agent powered by GPT-4o~\cite{openAI4o2024}, with its prompt detailed in Section~\ref{sec:AJ_prompt}.
For the given essay question, it is designed to determine which of the two answers is better based on principles of diversity, specificity, and professionalism.

\begin{equation}
J(q, a_1, a_2) =
\begin{cases}
1.0, & \small\text{$a_1$ is better than $a_2$} \\
0.0, & \small\text{$a_1$ is worse than $a_2$} \\
0.5, & \small\text{$a_1$ and $a_2$ are comparable}
\end{cases}
\label{equation:judging_agent}
\end{equation}

where $a_1$ and $a_2$ are two input answers of judging agent.


\section{Evaluation Prompt}
\label{sec:eval_prompt}

There are two types of prompts used in the evaluation process, the question answering (QA) prompt and the answer judging (AJ) prompt of essay question.
We introduce them in the following subsections, where
variables are represented in the format \$\{var\_name\}.

\subsection{Question Answering Prompt}
\label{sec:qa_prompt}

\tcbset{
    left*=10pt, right*=10pt,
    top=0pt, bottom=0pt,
    colback=white!10!white,
    colframe=black!75!black,
    fonttitle=\bfseries\large,
    subtitle style={boxrule=0pt,colback=gray!50!white},
    before=\vspace{-1em}, 
    after=\vspace{1em} 
}
\lstset{basicstyle=language=python, breaklines=true}
\begin{tcolorbox}[title= - QA Prompt]
\small
\setlength{\tabcolsep}{0.1mm}{
Image prompt:
\newline
\$\{selected sub-images in geologic map\}
\newline

Instruction prompt:
\newline
Extracted information:
\$\{information\}
\newline
Injected knowledge:
\$\{knowledge\}
\newline
This is a \$\{question type\} question.
\newline
Based on the provided text and image, reason and answer the question in JSON format only, for example: \{``reason": ``XXX", ``answer": ``XXX"\}
\newline

Question:
\newline
\$\{question\}
\newline
Answer: 
}
\end{tcolorbox}

\subsection{Answer Judging Prompt}
\label{sec:AJ_prompt}

\tcbset{
    left*=10pt, right*=10pt,
    top=0pt, bottom=0pt,
    colback=white!10!white,
    colframe=black!75!black,
    fonttitle=\bfseries\large,
    subtitle style={boxrule=0pt,colback=gray!50!white},
    before=\vspace{-1em}, 
    after=\vspace{1em} 
}
\lstset{basicstyle=language=python, breaklines=true}
\begin{tcolorbox}[title= - AJ Prompt]
\small
\setlength{\tabcolsep}{0.1mm}{
Image prompt:
\newline
\$\{entire image of geologic map\}
\newline

Instruction prompt:
\newline
Please evaluate which of the two answers below is better for the essay question \$\{question\}, consider the following criteria:
\newline
1. Diversity: The answer should address various aspects of the question, providing a well-rounded perspective.
\newline
2. Specificity: The answer should be detailed and precise, avoiding vague or general statements.
\newline
3. Professionalism: The answer should be articulated in a professional manner, demonstrating expertise and credibility.
\newline

Answer1:\
\newline
\$\{answer1\}
\newline
Answer2:\
\newline
\$\{answer2\}
\newline

Question: which answer is better?
\newline
A. Answer1 is better than Answer2
\newline
B. Answer1 is worse than Answer2
\newline
C. Answer1 and Answer2 are comparable
\newline

Only respond answer with A, B or C in JSON format, for example: \{``answer": ``C"\}
\newline
Answer: 
}
\end{tcolorbox}



\section{Evaluation Setting}
\label{sec:eval_setting}
\noindent\textbf{Base Model.}
We set all the random seeds to 42, the temperature to 0, and the maximum tokens to 2048 for base models.
Among them, we enable structured mode to enforce responses in JSON format for GPT-4o and GPT-4o-mini.
This functionality is not applied to other open-source MLLMs as they do not support it.
The system prompt is set to ``You are an expert in geology and cartography with a focus on geologic map.".

\noindent\textbf{Detection Model.}
\label{sec:det_model_set}
We use YOLOV10~\cite{wang2024yolov10} as the detection framework to train the map component detector and legend unit detector models.
The training settings are as follows:
input images are resized to 640$\times$640 for both detectors, SGD is employed as optimizer with initial learning rate of 0.01 and finnally linear decay to 0.0001.
The weight decay is set to 0.0005, and the total number of epochs is 500.
The models are trained on single GPU (80GB NVIDIA Ampere A100), where the batch size is 32.
We select and annotate approximately 1k original geologic maps as training dataset, ensuring no overlap with the GeoMap-Bench dataset.
During the inference stage, the Intersection over Union (IoU) threshold for Non-Maximum Suppression (NMS) is set to 0.8.

\noindent\textbf{GEE APIs.}
In Google Earth Engine (GEE)~\cite{gorelick2017google}, we use API of ``WorldPop/GP/100m/pop"~\cite{sorichetta2015high,gaughan2013high} image collection to retrieve population density data and API of ``ESA/WorldCover/v200"~\cite{zanaga2022esa} image collection for land cover data.
The scale for both collections is set to 100.

\noindent\textbf{Scientific DBs.}
We use USGS earthquake database~\cite{earthquake1977USGS} to retrieve records of historical earthquake data with magnitudes greater than 2.5 occurring since the 1970s.
For active faults database, we use GEM DB~\cite{styron2020gem}, which currently encompasses most of the deforming continental regions on Earth, with the exceptions of the Malay Archipelago, Madagascar, Canada, and a few other areas.


\section{Additional Experiment}
\label{sec:additional_exp}

\subsection{Overall Performance Comparison}

\begin{figure}[htbp]
  \centering
  \includegraphics[width=3.3in]{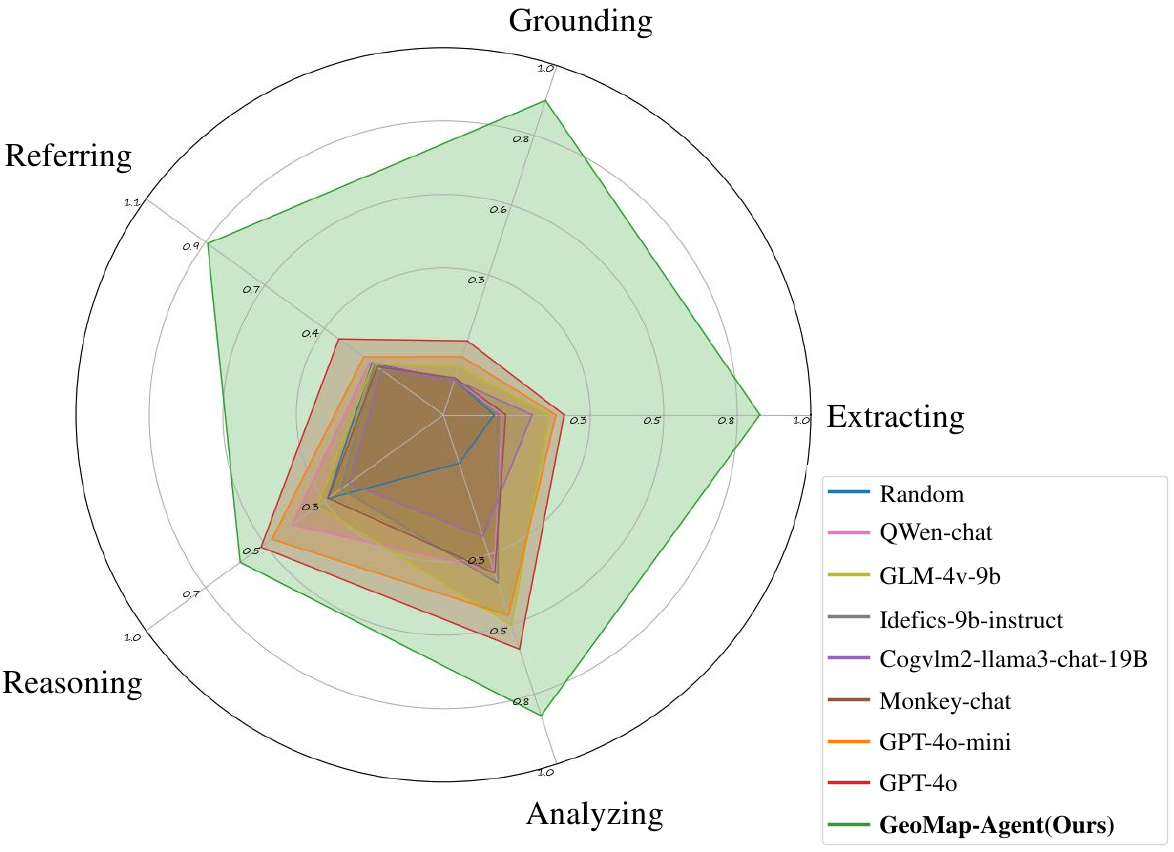}
  \hfill
\vspace{-10px}
  \caption{Overall performance comparison on different abilities.
  }
  \label{fig:per_comp_ability}
\end{figure}

\begin{figure}[htbp]
  \centering
  \includegraphics[width=3.3in]{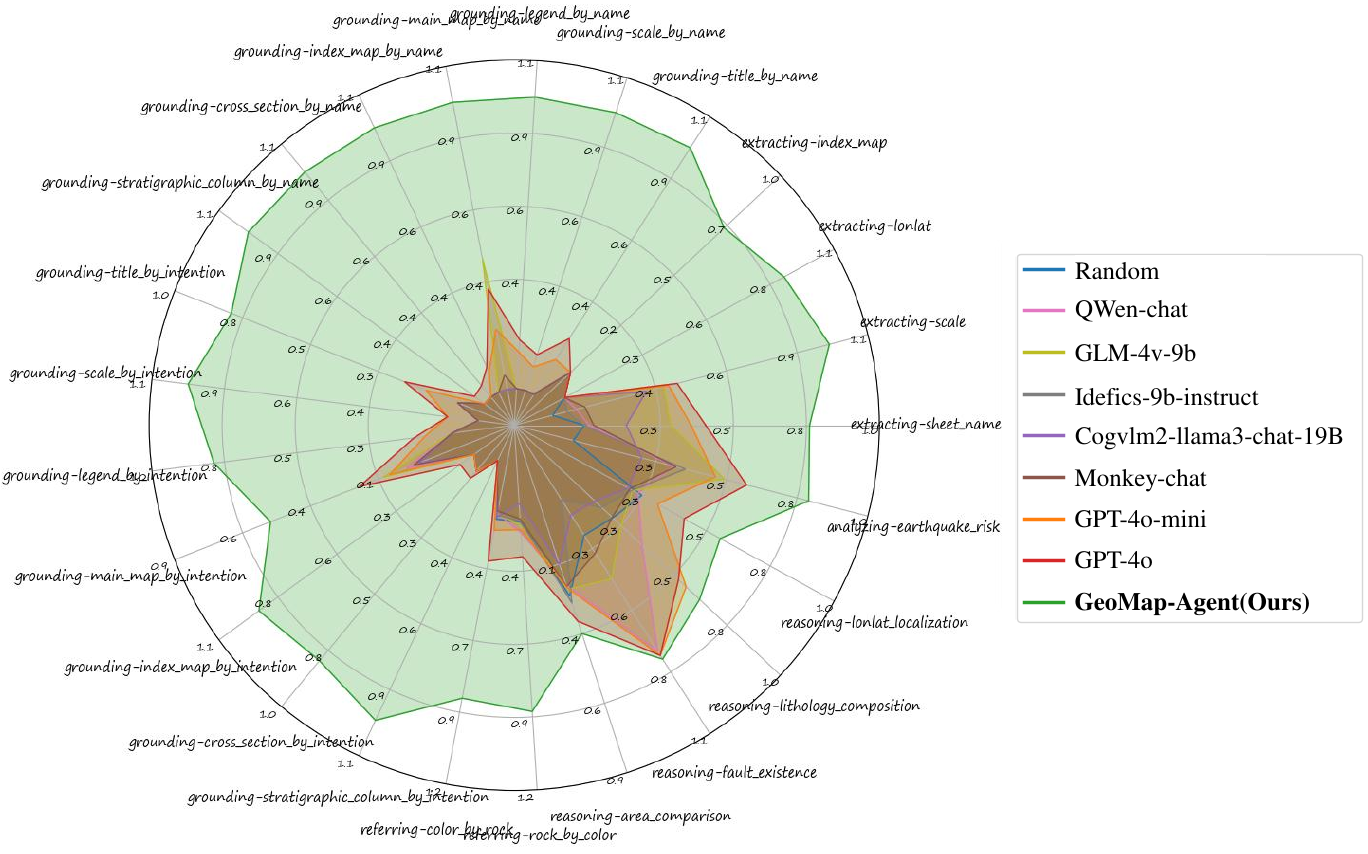}
  \hfill
\vspace{-10px}
  \caption{Overall performance comparison on different tasks.
  }
  \label{fig:per_comp_task}
\end{figure}

We compare the performance of different methods evaluated on the entire GeoMap-Bench dataset at both the ability and the task levels.
To visually present these results, we use radar charts, as shown in Figure~\ref{fig:per_comp_ability} and Figure~\ref{fig:per_comp_task}.
The results demonstrate that (1) GPT-4o is the best publicly available MLLMs on GeoMap-Bench across various abilities and tasks.
(2) Our method, GeoMap-Agent, significantly outperforms all the public MLLMs using GPT-4o as the base model.

\subsection{Improvement from Prompt Enhancement}
In the last PEQA module, we further improve GeoMap-Agent by enhancing its prompt from 4 aspects.
Aside from the first context enhancement, which relies on the global metadata and external knowledge from the initial two modules, the other three can be applied independently.
To evaluate the effectiveness, we conduct experiments with and without the last 3 enhancements in the PEQA module.
As shown in Table~\ref{tab:ablation_prompt}, these enhancements led to additional improvements on GeoMap-Bench.

\begin{table}[ht]
  \centering
  \begin{tabular}{@{}llcc@{}}
    \toprule
    Dataset & Ability & \makecell{enhance prompt \\ w/} & \makecell{enhance prompt \\ w/o} \\
    \hline
    \multirow{6}*{\makecell{USGS \\ Set}} & Ext. & \textbf{0.379} & 0.208 \\
    & Gro. & \textbf{0.123} & 0.100 \\
    & Ref. & \textbf{0.415} & 0.398 \\
    & Rea. & 0.491 & \textbf{0.494} \\
    & Ana. & \textbf{0.733} & 0.683 \\
    & Ove. & \textbf{0.428} &  0.376 \\
    \hline
    \multirow{6}*{\makecell{CGS \\ Set}} & Ext. & \textbf{0.326} & 0.230 \\
    & Gro. & \textbf{0.258} & 0.157 \\
    & Ref. & 0.331 & \textbf{0.359} \\
    & Rea. & \textbf{0.547} & 0.521 \\
    & Ana. & \textbf{0.584} & 0.542 \\
    & Ove. & \textbf{0.409} &  0.361 \\
    \bottomrule
  \end{tabular}
\vspace{-5px}
  \caption{Performance comparison of GeoMap-Agent on GeoMap-Bench with and without prompt enhancement. 
  All other settings remain the same, including the use of GPT-4o as base model and \textbf{excluding the HIE and DKI modules}.}
  \label{tab:ablation_prompt}
\vspace{-10px}
\end{table}


\section{Other Tools}
\label{sec:other_tools}

\subsection{Lithological Mapping Table}
To incorporate lithological knowledge into GeoMap-Agent, our professional geologists compile a 3-level lithological table (rock type, rock category, and lithology), containing 335 items in English and 256 items in Chinese, which is scalable as well.
A sample of the English lithological table is presented in Table~\ref{tab:lithological_table}.


\subsection{Legend Unit Extractor}
The legend unit is a standardized component across different geologic map sources.
We develop a tool for information extraction within each legend unit, encompassing both text and color extraction.
This process is based on the bounding box pairs of text unit and color unit detected by legend unit detector~\ref{sec:det_model_set}.
For text extraction, we employ the base model to process each cropped legend text unit, using the prompt ``Only output the OCR result of the given image.".
For color extraction, we calculate the median color in each cropped legend color unit.



\begin{table}[ht]
  \centering
  \begin{tabular}{@{}lcc@{}}
    \toprule
    
    Class & Subclass & Lithology \\
    \hline
    
    \multirow{6}*{Sedimentary} & \multirow{3}*{Clastic} & conglomerate \\
    \cline{3-3}
    &  & tillite \\
    \cline{3-3}
    &  & breccia \\
    \cline{2-3}
    & \multirow{3}*{Carbonate} & limestone \\
    \cline{3-3}
    &  & marl \\
    & ... & ... \\
    \hline
    
    \multirow{6}*{Volcanic} & \multirow{3}*{Acid volcanic} & trachydacite \\
    \cline{3-3}
    &  & keratophyre \\
    \cline{3-3}
    &  & quartz keratophyre \\
    \cline{2-3}
    & \multirow{3}*{Alkali volcanic} & analcimite \\
    \cline{3-3}
    &  & leucitite \\
    & ... & ... \\
    \hline

    \multirow{5}*{Intrusive} & \multirow{2}*{Acid intrusive} & tonalite \\
    \cline{3-3}
    &  & plagiogranite \\
    \cline{2-3}
    & \multirow{3}*{Alkaline intrusive} & foid diorite \\
    \cline{3-3}
    &  & foid gabbro \\
    & ... & ... \\
    \hline

    \multirow{7}*{Metamorphic} & \multirow{3}*{Slate} & siliceous slate \\
    \cline{3-3}
    &  & charcoal slate \\
    \cline{3-3}
    &  & sandy slate \\
    \cline{2-3}
    & \multirow{4}*{Schist} & graphitic schist \\
    \cline{3-3}
    &  & actionlite schist \\
    \cline{3-3}
    &  & amphibole schist \\
    & ... & ... \\

    \bottomrule
  \end{tabular}
  \caption{Sampled lithological mapping table.}
  \label{tab:lithological_table}
\end{table}


\newpage

{
    \small
    \bibliographystyle{ieeenat_fullname}
    \bibliography{main}
}


\end{document}